\documentclass{article}
\usepackage{style}

\usepackage{times}
\usepackage{soul}
\usepackage{url}
\usepackage[hidelinks]{hyperref}
\usepackage[utf8]{inputenc}
\usepackage[small]{caption}
\usepackage{graphicx}
\usepackage{amsmath}
\usepackage{amsthm}
\usepackage{booktabs}
\usepackage{algorithm}
\usepackage{algorithmic}
\usepackage[switch]{lineno}

\usepackage{amssymb}
\usepackage{dsfont}

\newtheorem{thm}{Theorem}
\newtheorem{lem}{Lemma}
\newtheorem{ass}{Assumption}
\newtheorem{theorem}[thm]{Theorem}
\newtheorem{lemma}[lem]{Lemma}
\newtheorem{assumption}[ass]{Assumption}

\title{LocMoE: A Low-Overhead MoE for Large Language Model Training}

\author{
	Jing Li\footnotemark[2]
	\and
	Zhijie Sun\footnotemark[2]$^,$\thanks{Corresponding author}
	\and
	Xuan He\footnotemark[2]
	\and
	Li Zeng
	\and
	Yi Lin
	\and
	Entong Li
	\and
	Binfan Zheng
	\and
	Rongqian Zhao
	\And
	Xin Chen\\
	\affiliations
	Huawei Technologies Co., Ltd\\
	\emails
	\{lijing473, sunzhijie3, hexuan22, zengli43, linyi11, lientong, zhengbinfan1, zhaorongqian, chenxin\}@huawei.com
}

\begin{document}

\maketitle

\begin{abstract}
The Mixtures-of-Experts (MoE) model is a widespread distributed and integrated learning method for large language models (LLM), which is favored due to its ability to sparsify and expand models efficiently. However, the performance of MoE is limited by load imbalance and high latency of All-to-All communication, along with relatively redundant computation owing to large expert capacity. Load imbalance may result from existing routing policies that consistently tend to select certain experts. The frequent inter-node communication in the All-to-All procedure also significantly prolongs the training time. To alleviate the above performance problems, we propose a novel routing strategy that combines load balance and locality by converting partial inter-node communication to that of intra-node. Notably, we elucidate that there is a minimum threshold for expert capacity, calculated through the maximal angular deviation between the gating weights of the experts and the assigned tokens. We port these modifications on the PanGu-$\Sigma$ model based on the MindSpore framework with multi-level routing and conduct experiments on Ascend clusters. The experiment results demonstrate that the proposed LocMoE reduces training time per epoch by 12.68\% to 22.24\% compared to classical routers, such as hash router and switch router, without impacting the model accuracy.
\end{abstract}

\section{Introduction}

Large Language Models (LLM), such as GPT \cite{brown2020language} and LLaMA \cite{touvron2023llama}, have recently gone viral due to their distinguished capabilities in word processing and data analysis. The architectures of these LLMs are mostly derived from the Transformer, which is on the basis of the self-attention mechanism \cite{vaswani2017attention}. Since the predictive ability of the Transformer-based model correlated strongly with the model size \cite{kenton2019bert}, the parameter scales of existing LLMs have increased dramatically to assure accuracy. The complex construction, along with the large parameter scale, triggers the rapid surge in demand for computing resources, resulting in escalating training and inference costs that hinder the development of LLMs \cite{lewis2021base}. Aiming at the problem, Mixtures-of-Experts (MoE) \cite{jacobs1991adaptive} provide an effective way to extend the model capacity at a fixed computational overhead \cite{he2021fastmoe}, thus emerging as the preferred option for some renowned LLMs.

A typical MoE framework consists of a gated network and several expert networks that selectively activate a portion of parameters for various inputs to participate in computation \cite{clark2022unified}. Owing to such a structure, the computational complexity remains relatively invariant when the scale of parameters increases \cite{puigcerver2020scalable}. Since each token activates only one or a few experts, sparse routing of the gated network delivers the token to the most appropriate expert(s) \cite{zuo2021taming}. If the routing strategy is not well-designed, it may lead to the overtraining of a few experts and under-training of others, ultimately evolving into inefficient learning and uneven load distribution \cite{shazeer2016outrageously}. To address this shortcoming, Switch Transformer \cite{fedus2022switch} simplifies the routing mechanism of MoE while adding an auxiliary loss that encourages a balanced load across experts. Moreover, the frequent All-to-All communication delay has also limited the performance of MoE \cite{rajbhandari2022deepspeed}. It is estimated that the time-consuming ratio of All-to-All under 8 A100 GPUs in a single node is about 31.18\% and would be much higher in multiple nodes \cite{nie2023flexmoe}. HeTuMoE \cite{nie2022hetumoe} further puts forward a hierarchical All-to-All strategy, which fully utilizes the bandwidth of intra-node NVLink and inter-node Infiniband to cope with the problem of low bandwidth utilization due to frequent inter-machine transfers of small data volumes.

In this paper, we propose \textbf{LocMoE}, a low-overhead routing strategy and a communication optimization scheme, and it is applied in PanGu-$\Sigma$ model \cite{ren2023pangu}. PanGu-$\Sigma$ is a sparse model extended by the dense model PanGu-$\alpha$ \cite{zeng2021pangu}. With Ascend cluster \cite{liao2021ascend}, it is measured that the All-to-All communication in PanGu-$\Sigma$ takes 18.10\% and 28.74\% of the training time under 128 Ascend 910A Neural Network Processing Units (NPUs) and 256 Ascend 910A NPUs, respectively. It still has potential for further reduction, and we make the following optimizations based on: 

\begin{itemize}
	\item \textbf{Orthogonal gating weight with Grouped Average Pooling (GrAP) layer.} The GrAP layer is adopted in gating values computation. It provides a natural way to perform class activation mapping and reduce computational costs. Above all, the orthogonality of gating weight facilitates the explicit decisions of the router.
	
	\item \textbf{Locality-based expert regularization.} Redistribute on the basis of load balance, add the locality loss as the regularization term, and transform partial inter-node communication into intra-node communication with higher bandwidth. The local experts are encouraged to compete with skilled experts, and the time consumption of communication is reduced while avoiding the under-training of some experts.
	
	\item \textbf{Reduction of expert capacity without losing accuracy.} Our work proves and solves the critical value of MoE's expert capacity in the NLP sector for the first time, and its relationship with input corpus features is also elucidated. Furthermore, we find fewer class-discriminative tokens need to be learned by experts than class-correlated ones. The experimental results also confirm that the model accuracy would not be affected after downward adjusting the expert capacity within the critical limit. 
\end{itemize}

After applying the above improvements, the time consumption of All-to-All communication decreases by 5.13\%. The elapsed time per epoch decreases by up to 22.24\% with our cluster groups (containing 8, 16, and 32 node with 64, 128, and 256 Ascend 910A NPUs, abbreviated as 64N, 128N, and 256N in the following paragraphs).

The remainder of this paper is organized as follows: Section 2 displays related works of MoE in the field of NLP, the Ascend architecture, and the base model PanGu-$\Sigma$. Section 3 demonstrates the methodology details of the LocMoE and the theoretical bounds. Section 4 analyses the results of comparison experiments. Section 5 summarizes this work and the prospects for its future research orientation.

\section{Related Work}

\paragraph{MoE.} MoE is a strategy for model designing, combining with several expert networks, to enhance the model capacity and efficiency. The concept of MoE was first proposed in 1991 and became the prototype of the existing MoE structure \cite{jacobs1991adaptive}. Sparsely-gated MoE \cite{shazeer2016outrageously} was proposed to expand the model capacity adequately under the same arithmetic power, and the gating is designed to allow TopK experts to be activated in an iteration. GShard \cite{lepikhin2020gshard} was the first work to migrate the MoE to Transformer, using the expert capacity to limit the tokens processed by each expert to a certain range. In addition, the auxiliary loss is proposed in GShard's random routing to deal with the winner-take-all drawback of MoE. Regarding expert capacity, the work of pMoE \cite{chowdhury2023patch} has proved for the first time that each expert can be fully trained even when dealing with samples much smaller than the number of tokens, but has a threshold. Switch Transformer \cite{fedus2022switch} selects only the top expert to maximize MoE's sparsity and proposes a corresponding auxiliary loss to achieve load balance. Facebook AI Research implements the Hash FFN layer \cite{roller2021hash} with the balanced hash function, and the distribution of the experts' load is close to the ideal state. Taking into consideration both convergence and accuracy, StableMoE \cite{dai2022stablemoe} adopts a two-stage training procedure. In the first stage, the imbalance of assignment and the cross-entropy of routing features are adopted as loss penalty terms, and the model directly learns with the routing strategy in the second stage. X-MoE \cite{chi2022representation} rewrites the score function between the token and the expert by reducing dimensionality. Task-MoE \cite{kudugunta2021beyond} describes task-based routing at multiple granularities: token level, sentence level, and task level. HetuMoE \cite{nie2022hetumoe} proposes the hierarchical AlltoAll strategy, which combines hierarchical networks and aggregated information to improve transmission efficiency.

\begin{figure}
	\centerline{\includegraphics[width=6cm]{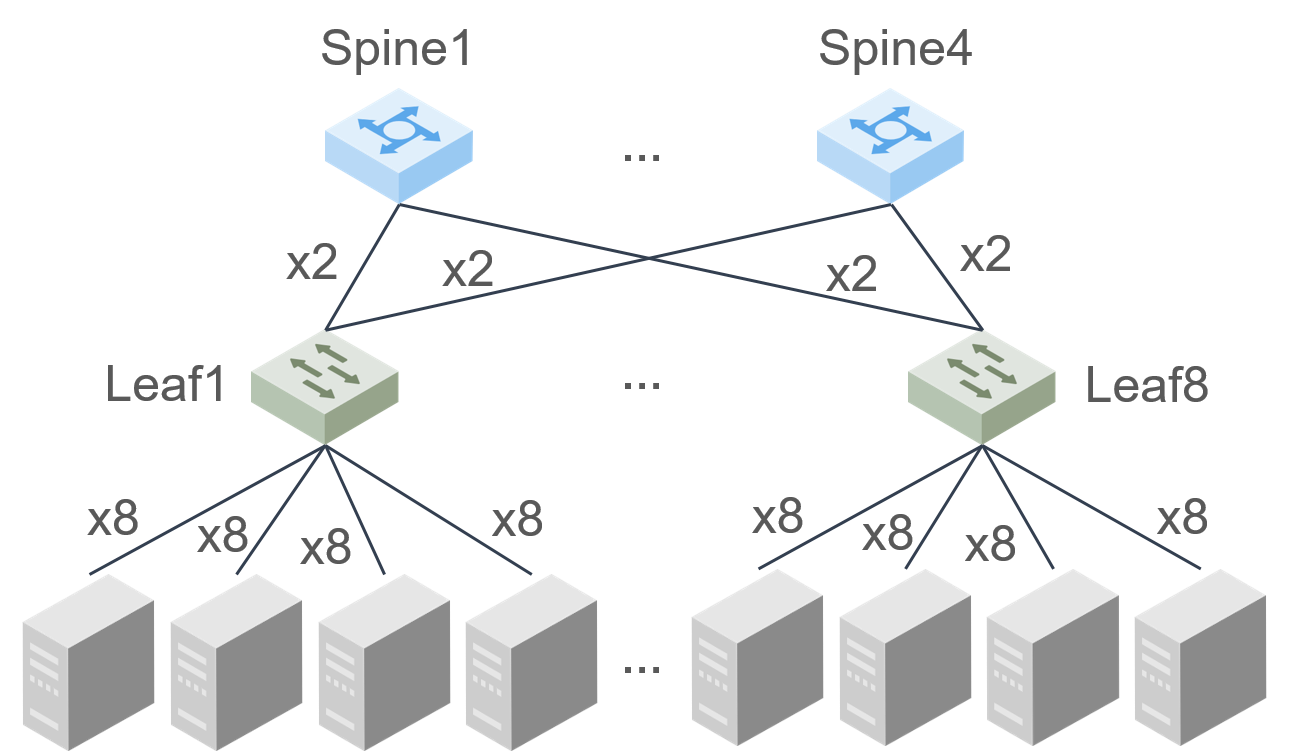}}
	\caption{The networking scheme applied in the Ascend cluster.}
	\label{network}
\end{figure}

\begin{figure}
	\centerline{\includegraphics[width=8cm]{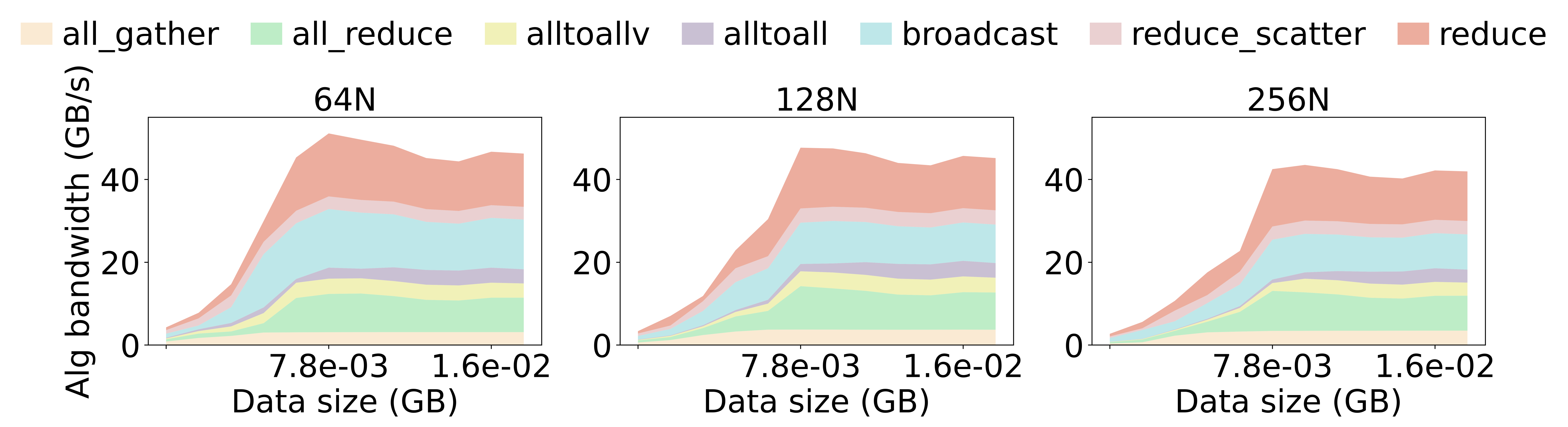}}
	\caption{The algorithm bandwidth of each communication operator in HCCL under 64N, 128N, and 256N, respectively.}
	\label{hccl}
\end{figure}

\paragraph{Ascend Architecture.} The pivot architecture of the Ascend mainly consists of multilevel on-chip memory, load/storage units, and instruction management units \cite{liao2021ascend}. System-on-Chip (Soc) adopts the Mesh Network-on-Chip (NoC) \cite{kumar2022architecture} architecture to provide a unified and scalable communication network, realizing a high bandwidth of 256GB/s \cite{li2022ascend}. In Ascend 910A server, every eight NPUs are divided into two groups on the board. The intra-group connection is based on the Huawei Cache Coherence System (HCCS) \cite{xia2021kunpeng}. The Ascend 910A chip delivers 320 Tera FLOPS at semi-precision (FP16) and 640 Tera OPS at integer precision (INT8). Our cluster is built based on a two-tier Fat-tree networking scheme on the single plane, with each Leaf switch connecting to 4 NPU servers (model Atlas 800 9000), as in Figure \ref{network}. The algorithm bandwidth of each communication operator in Huawei Collective Communication Library (HCCL) is displayed in Figure \ref{hccl}.

\paragraph{PanGu Series Model.} The fields of PanGu series large models are mainly divided into NLP, computer vision, multimodality, graph network, and scientific computing \cite{mi2022pangu,shen2023pangu,bi2023accurate}. Thereinto, the models in the field of NLP focus primarily on text generation and semantic understanding. The most representative NLP model in the PanGu series is the PanGu-$\alpha$ \cite{zeng2021pangu}, which is an LLM in the Chinese domain with up to 200 billion parameters. It also applies the auto-parallel framework based on the MindSpore \cite{tong2021study}. PanGu-$\pi$ \cite{wang2023pangu} mitigates feature collapse in the Transformer architecture by introducing more nonlinearities in the feed-forward networks (FFN) and MSA modules. Utilizing the intrinsic parameters of PanGu-$\alpha$, PanGu-$\Sigma$ \cite{ren2023pangu} is extended to a sparse model containing 1.085 trillion parameters by the conception of MoE.

\section{Methodology}

\subsection{PanGu-$\Sigma$}

\begin{figure}
	\centerline{\includegraphics[width=8cm]{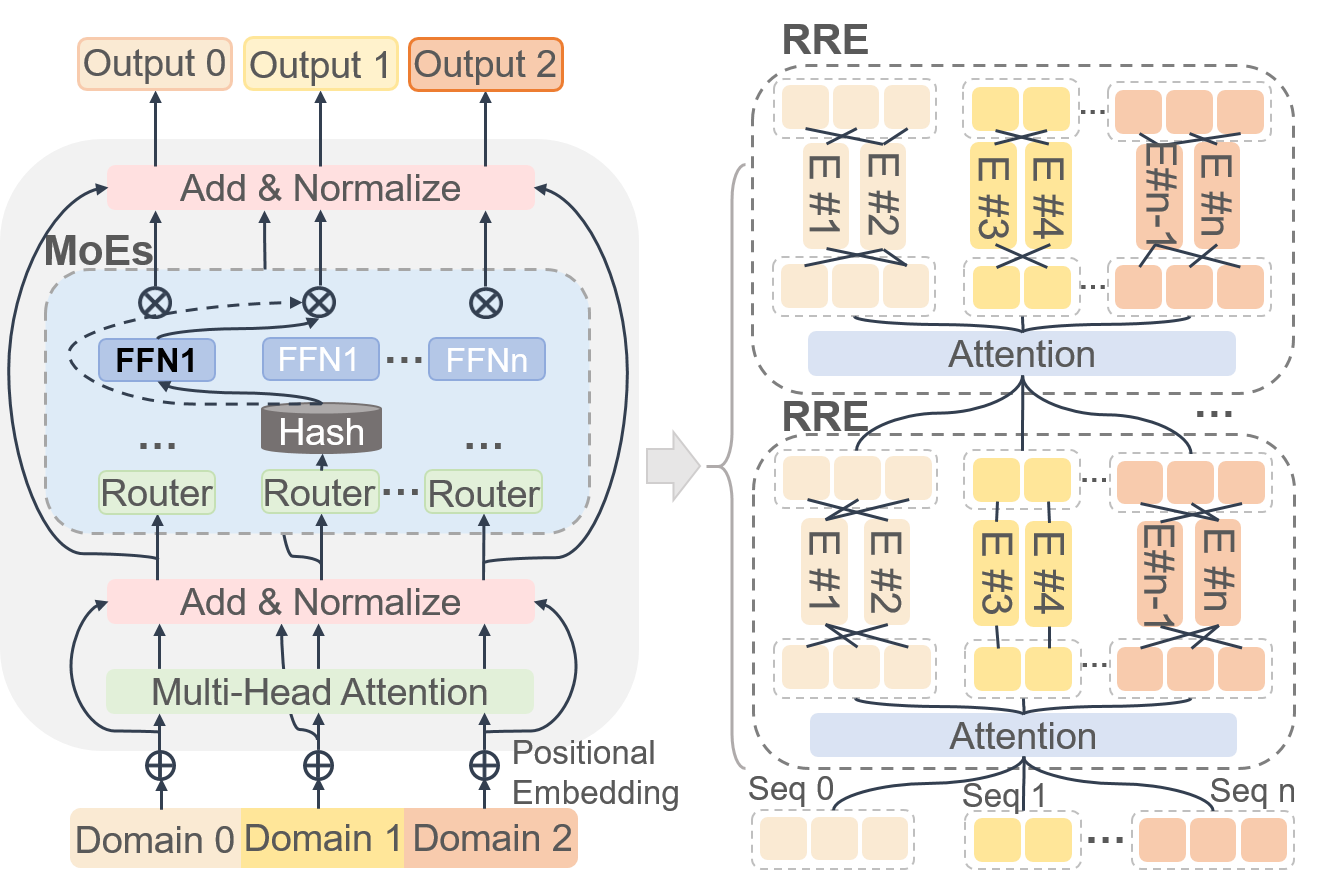}}
	\caption{The architecture of sparse Transformer layers in PanGu-$\Sigma$.}
	\label{pangu_sigma}
\end{figure}

The PanGu-$\Sigma$ architecture consists of both dense and sparse Transformer encoder layers, stacked Transformer decoder layers modeled in the autoregressive language, and a query layer. The sparse Transformer layer of PanGu-$\Sigma$, with several conditionally activated feedforward sublayers, incorporates the MoE principle, as displayed in Figure \ref{pangu_sigma}. The RRE module is responsible for routing the token to the appropriate expert. It contains two levels of routing: in the first level, the experts are grouped by domains, and the token is assigned to one of the groups. In the second level, the token is routed to a particular expert of this group homogeneously. The second level of routing can be viewed as random hash routing, which does not contain learnable parameters.

\subsection{MoE With Local Routing Strategy}

\subsubsection{MoE in Encoder Layers of Transformers}

Similar to the classic MoE skeleton applied to Transformer structures such as GShard, the MoE layer in our model mainly consists of a MSA layer, a gating network, a routing module, and several expert FFNs. The output of the MoE layer can be depicted as follows:

\begin{equation}
	y_m = \sum_{i=1}^{n}\mathcal{R}_{m, E_i}\cdot{W_{E_i, \mathrm{out}}\cdot{\mathrm{GeLU}(W_{E_i, \mathrm{in}}\cdot{x_m})}}
\end{equation}

Assume that the MoE layer contains $n$ experts, $\mathcal{R}_{m, E_i}$ denotes the expert score acquired by the gating network when expert $i$ provides the largest gating value. The expert network of expert $i$ consists of two linear transformations with a Gaussian Error Linear Unit (GeLU) activation, which is the product of input and the standard Gaussian cumulative distribution function. Thereinto, the gating function $\mathcal{G}$ is the critical component of router $\mathcal{R}$. Typically, it is designed to be a dense layer extracting the feature of input tensor:

\begin{equation}
	i^* = \mathop{\arg\max}\limits_{i\in{[n]}}(\mathrm{softmax}(\mathcal{G}_{m, E_i}))
	\label{equ:max_expert}
\end{equation}

\begin{equation}
	\mathcal{R}_{m, E_i} = \mathds{1}\{i=i^*\}(\mathrm{softmax}(\mathcal{G}_{m, E_i}))
	\label{equ:gating_score}
\end{equation}

\begin{figure*}
	\centerline{\includegraphics[width=16cm]{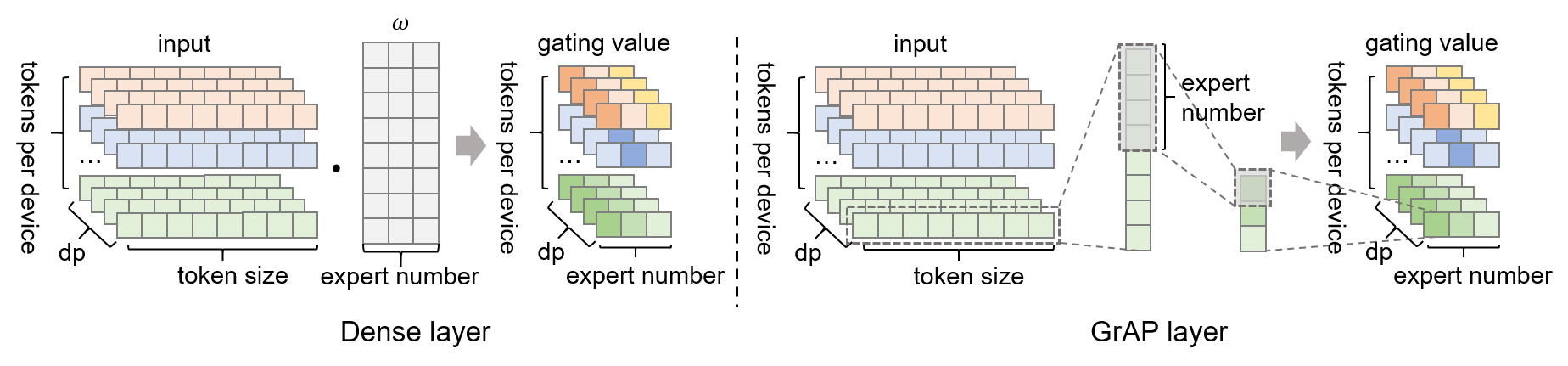}}
	\caption{Difference between feature extraction via the dense layer and the GrAP layer.}
	\label{gap}
\end{figure*}

where $i^*$ is the index of the most appropriate expert, and $\mathcal{G}_{m, E_i} = \mathrm{ReLU}(\omega_i \cdot x_m + \epsilon_i)$. The token would be sent to the Top-1 expert with the largest expert score screened by Softmax. To reduce the parameter scale and the computational overhead, the gating value is obtained via the GrAP layer instead of the dense layer \cite{li2022multi}. The feature extraction with the GrAP layer is delineated in Figure \ref{gap}. It can be regarded as the dense layer with the fixed weight $\omega_i$:

\begin{equation}
	\omega_i = \frac{n}{d}(\omega_{i, j} = \mathds{1}\{i\frac{d}{n}\leq{j}<{(i + 1)\frac{d}{n}}\}, 0\leq{j}<d)
\end{equation}

where $d$ denotes the dimension of activation. Notably, the gating weights of the GrAP layer are orthogonal. From a perspective of semantic, irrelevant tokens are inclined to be routed to experts of different domains, which is conducive to convergence and accuracy \cite{guo2020real}. Besides, the GrAP layer has greater computation efficiency.

\subsubsection{Localized Bias Weighting Loss}

A general observation on the original two-level routing strategy of PanGu-$\Sigma$ reveals that the router is devoid of the learning process. Although meeting load balance requirements, it lacks interpretability for distinguishing experts by domain. LocMoE rewrites the second level of RRE, consisting of two parts: auxiliary loss and locality loss. The auxiliary loss is first proposed in the sparsely-gated MoE \cite{shazeer2016outrageously} and is also applied in Switch Transformer \cite{fedus2022switch}:

\begin{equation}
	\begin{aligned}
		L_{\mathrm{aux}} &= \alpha{{n}{\sum_{i=1}^{n}f_i{P_i}}} \\
		f_i = \frac{1}{T}\sum_{x\in{\beta}}\mathds{1}\{\mathop{\arg\max}\,&p(x)=i\}, P_i = \frac{1}{T}\sum_{x\in{\beta}}p_i(x)
	\end{aligned}
\end{equation}

where $n$ denotes the number of experts, $\beta$ denotes the batch containing $T$ tokens. $f_i$ denotes the proportion of tokens assigned to expert $i$, and $P_i$ denotes the average probability that the router chooses expert $i$. The auxiliary loss has substantiated that it can cause the balance of routing, as the loss would achieve its minimum under a uniform distribution. The hyperparameter $\alpha$ is set to 0.01, which refers to the value in the previous work \cite{fedus2022switch}.

\begin{figure}
	\centerline{\includegraphics[width=5.5cm]{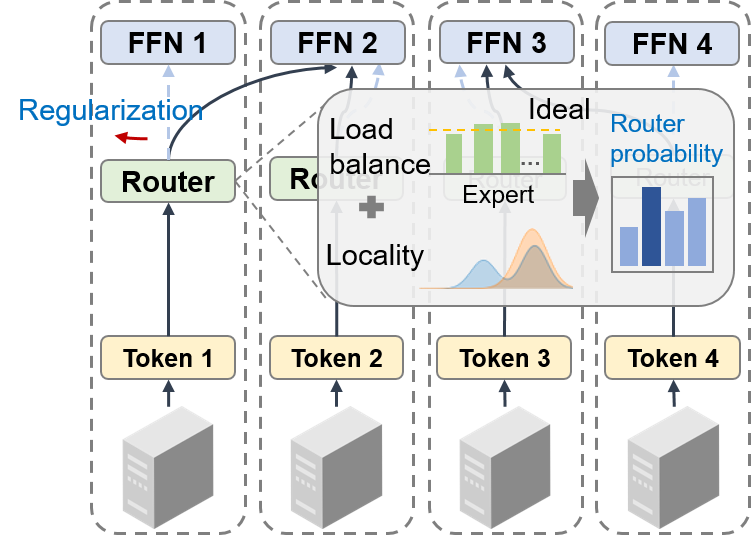}}
	\caption{The action principle of locality loss.}
	\label{locality}
\end{figure}

The second part is locality loss, in line with the expectation that tokens are more likely to be assigned to local experts under the premise of load balance. The loss function can be measured by the difference between the current distribution and the fully localized distribution. The current distribution reflects the assignment distribution of all experts in the current batch, and the difference can be described using Kullback-Leibler (KL) divergence:

\begin{equation}
	L_{\mathrm{loc}} = \mu \mathrm{KL}(D_{\mathrm{c}}||D_{\mathrm{l}}) = -\mu \int{D_{\mathrm{c}}(x)\ln[\frac{D_{\mathrm{l}}(x)}{D_{\mathrm{c}}(x)}]\mathrm{d}x}
\end{equation}

where $D_c$ denotes the current distribution and $D_l$ denotes the fully localized distribution, and $\mu$ is the hyperparameter. The locality, along with the auxiliary loss, acts as the soft constraint that impels the tokens in the same domain to be trained by local experts, as shown in Figure \ref{locality}. The blue dashed arrows are the contribution of the locality loss, and the final assignment of tokens considers the synthetic effect of gating value, auxiliary loss, and locality loss. The task loss is the sum of the above loss items and cross-entropy:

\begin{equation}
	\begin{aligned}
		L_{\mathrm{task}} &= L_{\mathrm{aux}} + L_{\mathrm{loc}} + L_{\mathrm{cross}} \\
		\mbox{where}\;L_{\mathrm{cross}} &= -\sum_{t=1}^T\log\frac{\exp(c_{t}^*)}{\sum_{i=1}^N\exp(c_{t, i})}
	\end{aligned}
\end{equation}

\subsubsection{Critical Value of Expert Capacity}

The introduction of expert capacity aims to avoid the training block caused by the assignment imbalance of tokens. In general, an empirical expert capacity factor $c_\mathrm{f}$ is set to limit the scale of the expert capacity: $ec = \lceil{\frac{b_{\mathrm{s}} * c_{\mathrm{f}}}{ep * n}}\rceil$. $b_\mathrm{s}$, $c_\mathrm{f}$, $ep$, and $n$ denote the batch size, the capacity factor, the degree of expert parallelism, and the number of experts, respectively. The computational workload of experts can be equalized in this way. However, the work of pMoE reveals that the sample size each expert needs to process has its lower bound \cite{chowdhury2023patch}, which can provably reduce the training costs. Inspired by the work on pMoE, we migrated the assumptions of data distribution in MoE from CV to the NLP domain, in conjunction with the network structure, while further discovering some significant conclusions:

\begin{assumption}\label{lma:gate} 
	The magnitudes of gating weight $\Vert\omega\Vert$ are equivalent for all experts.
\end{assumption}

\begin{lemma}[Minimum Angle of Expert]\label{lma:router} 
	The top-1 router is essentially the mechanism to select the expert $i^*$ with the minimum angle $\theta_{i^*, j}$ to the gating weight $\omega_i$.
\end{lemma}

\begin{assumption} Suppose all tokens with size of $d$ uniformly distributed on the unit sphere, that is, $\Vert{x_m}\Vert=1$.
\end{assumption}

\begin{lemma}[Equivalent Probability for Assignment]\label{lma:probability} 
	Suppose $i_j$ denotes the expert to which the token $j$ is routed. On account of the spherical symmetry, the probabilities for $i_j = i^{'}$ are equivalent for all experts under the conditions of orthogonal gating weights (brought from the GrAP layer). That is, $P\{i_j=i^{'}\}=\frac{1}{n}$.
\end{lemma}

\begin{assumption} Assume that if $\delta_{i, j}\geq{\delta}$, the token $j$ should be assigned to the expert $i$, where $\delta_{i, j} = \mathrm{cos}(\theta_{i, j})$
\end{assumption}

\begin{lemma}[Assignment Probability for Unit Vector]\label{lma:uniformly}
	For the uniformly distributed unit vector $j$, the probablity it should be assigned to the expert $i$ is:
	
	\begin{equation}
		\begin{aligned}
			p_{\delta} &= 1 - I_{\delta^2}(\frac{1}{2}, \frac{d-1}{2}) \\
			\mathrm{where}\;I_{x}(a, b) &= \frac{1}{B(a, b)}\int_{0}^{x}t^{a - 1}(1 - t)^{b - 1}\mathrm{d}t, \\
			0&\leq{x}\leq{1}, a>0, b>0
		\end{aligned}
		\label{thm:delta}
	\end{equation}
\end{lemma}

As for the activation size, for large $d$, when $\delta=\Theta{(\frac{1}{\sqrt{d}})}, p_\delta{\approx}0.3$. When $\delta$ is larger than $\frac{1}{\sqrt{d}}$, $p_\delta$ declines to $0$ rapidly.

\begin{figure}
	\centerline{\includegraphics[width=8cm]{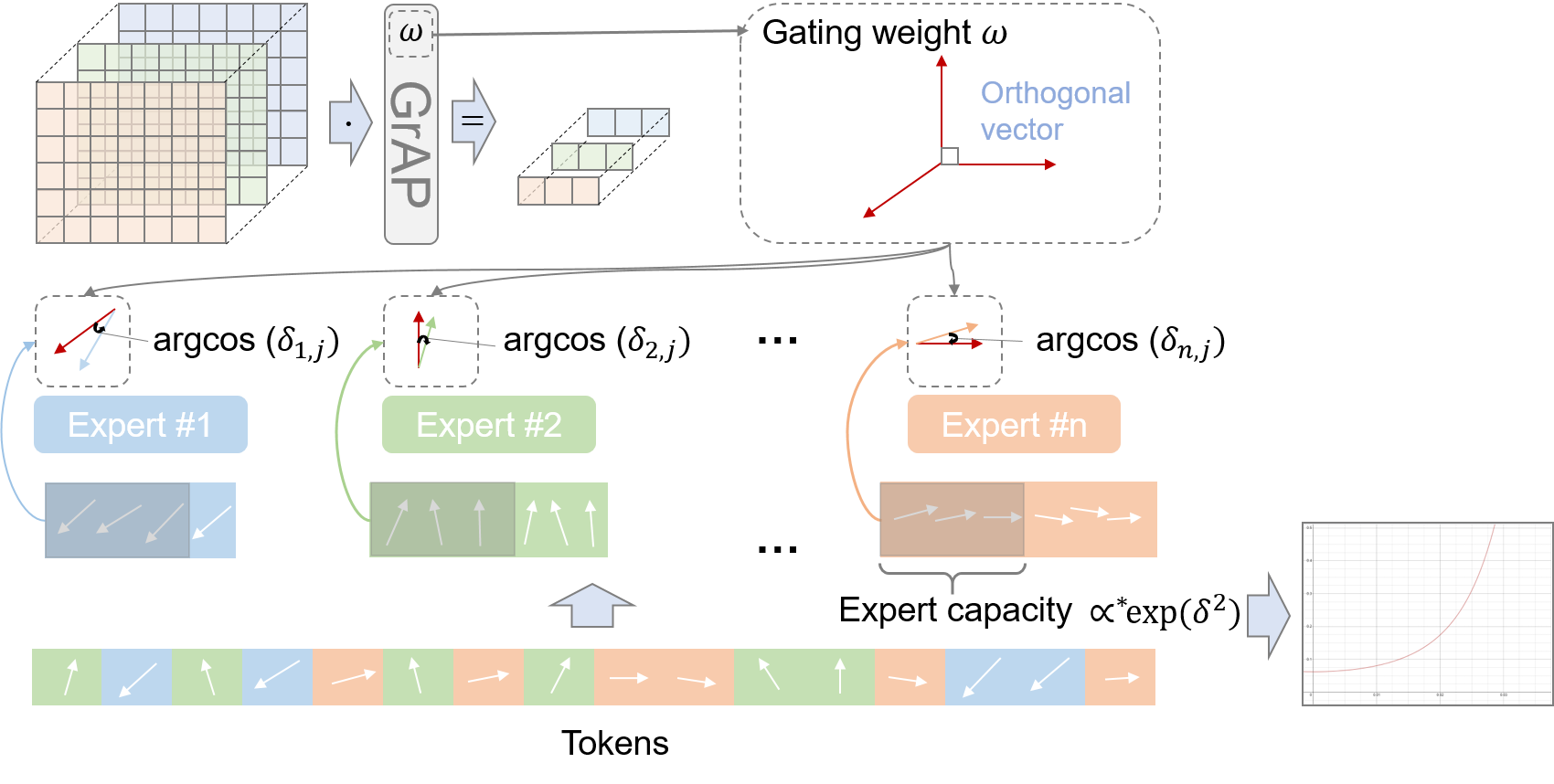}}
	\caption{The correlation between expert capacity and the angle between token and gating weight.}
	\label{expert_capacity}
\end{figure}

\begin{theorem}[Lower Bound of Expert Capacity]\label{lma:lower}
	From Lemma \ref{lma:uniformly}, assume that $p_i$ denotes the probability of the token routed to the expert $i$ is class-discriminative, thus, $p_i\leq{n[1 - I_{\delta^2}(\frac{1}{2}, \frac{d-1}{2})]}$. The lower bound of the expert capacity can be described as:
	
	\begin{equation}
		\begin{aligned}
			ec_{\min} &= \frac{1}{p_i} \geq \frac{1}{n[1 - I_{\delta^2}(\frac{1}{2}, \frac{d-1}{2})]} \\
			\mathrm{for\;large\;}d, \\
			ec_{\min} &\geq{\frac{1}{n\cdot{\mathrm{erfc}(\sqrt{\frac{\delta^2d}{2-\delta^2}})}}}>\frac{1}{n}\exp(\frac{\delta^2d}{2-\delta^2}),
		\end{aligned}
	\end{equation}
	
\end{theorem}

where erfc is the complementary error function. Figure \ref{expert_capacity} portrays the schematic diagram of our discovery, which describes the correlation between the expert capacity and the minimum angle of experts. The expert capacity correlates negatively with the minimum angle between token and gating weight, and it grows exponentially with the decrease of the angle $\theta$.

\subsubsection{Group-Wise All-to-All and Communication Overlap}

Since the All-to-All is an aggregate communication operator, other operations would be performed until the data is completely transmitted, leading to low hardware utilization efficiency. Our model applies the group-wise exchange algorithm embedded in MindSpore to split and rearrange the All-to-All operations. In the tensor parallel (TP) domain, each device is responsible for a portion of the All-to-All data transmission in its respective expert parallel (EP) domain. Then, the All-Gather operation is conducted to synchronize tokens on all devices in the TP domain. The communication volume is diverted to the TP domain with high-speed bandwidth, which reduces the overall All-to-All communication time. In addition, FFN computation and communication are sliced and overlapped to mask the delay caused by communication, eventually reducing the time of communication.

\section{Experiment Results and Analysis}

\begin{figure}
	\centerline{\includegraphics[width=6cm]{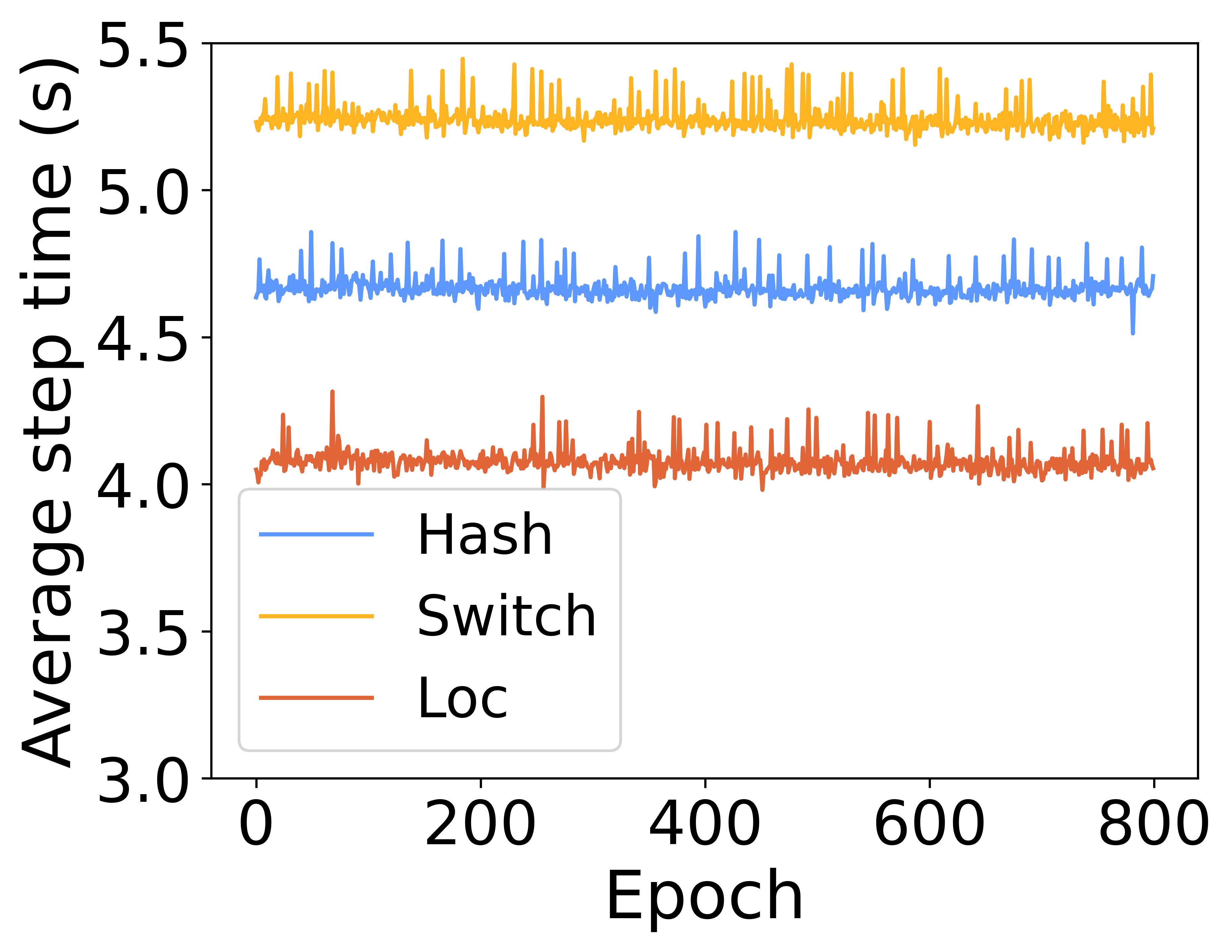}}
	\caption{The average time consumption of steps in each epoch with multiple MoEs under 64N.}
	\label{step_time}
\end{figure}

We conduct experiments on the Ascend cluster groups (see environment configuration in Appendix C) to verify the effect of LocMoE. The existing classical MoEs, such as HashMoE and SwitchMoE, are implemented in PanGu-$\Sigma$ and made contrasts. The average training time with these MoEs under 64N is displayed in Figure \ref{step_time}. It can be seen that LocMoE has an average speedup of 1.15$\times$ to 1.29$\times$ compared to HashMoE and SwithMoE, respectively.

\subsection{Analysis for Expert Capacity}

\begin{figure}
	\centerline{\includegraphics[width=8cm]{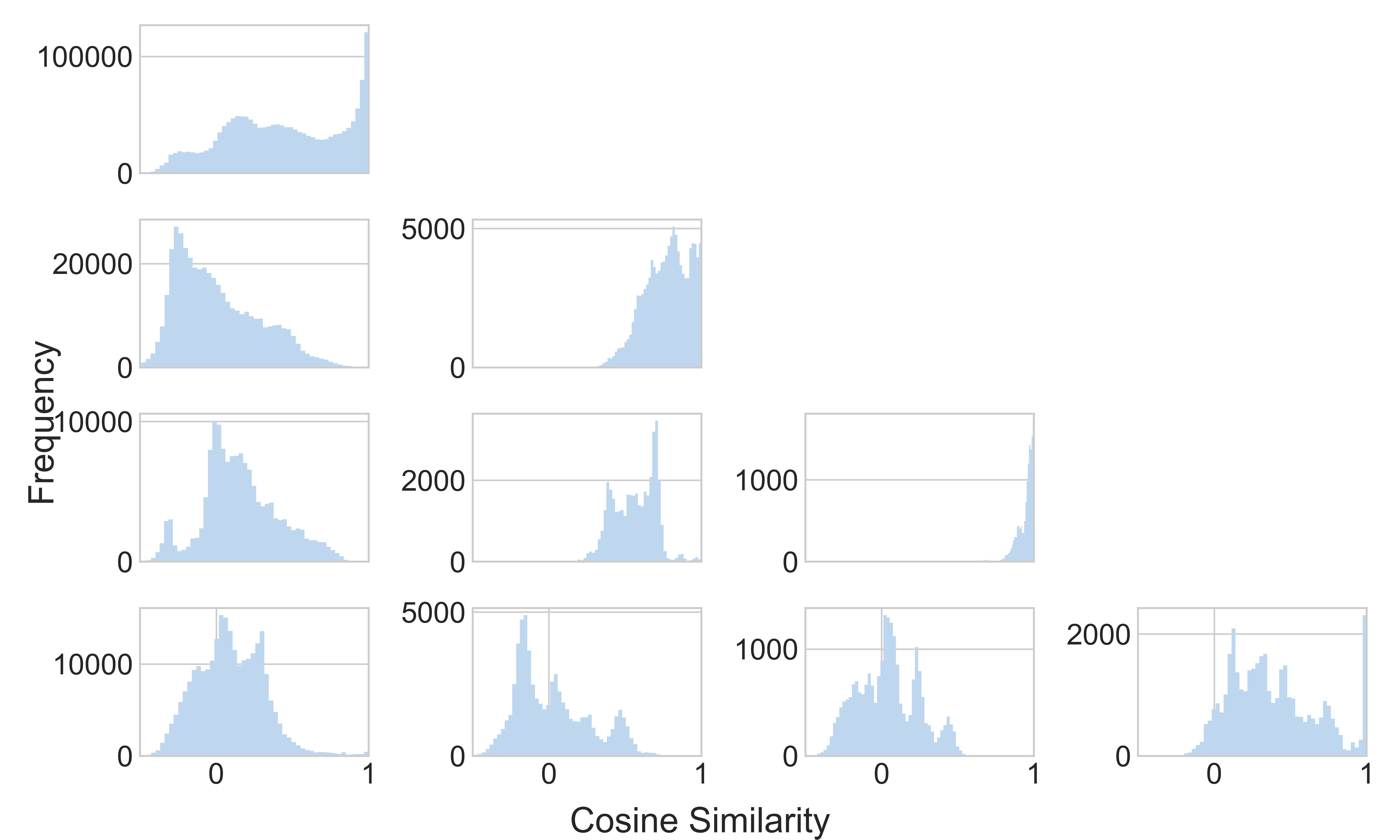}}
	\caption{The histograms of cosine similarities between tokens routed to two experts.}
	\label{freq_1}
\end{figure}

\begin{figure}
	\centerline{\includegraphics[width=7cm]{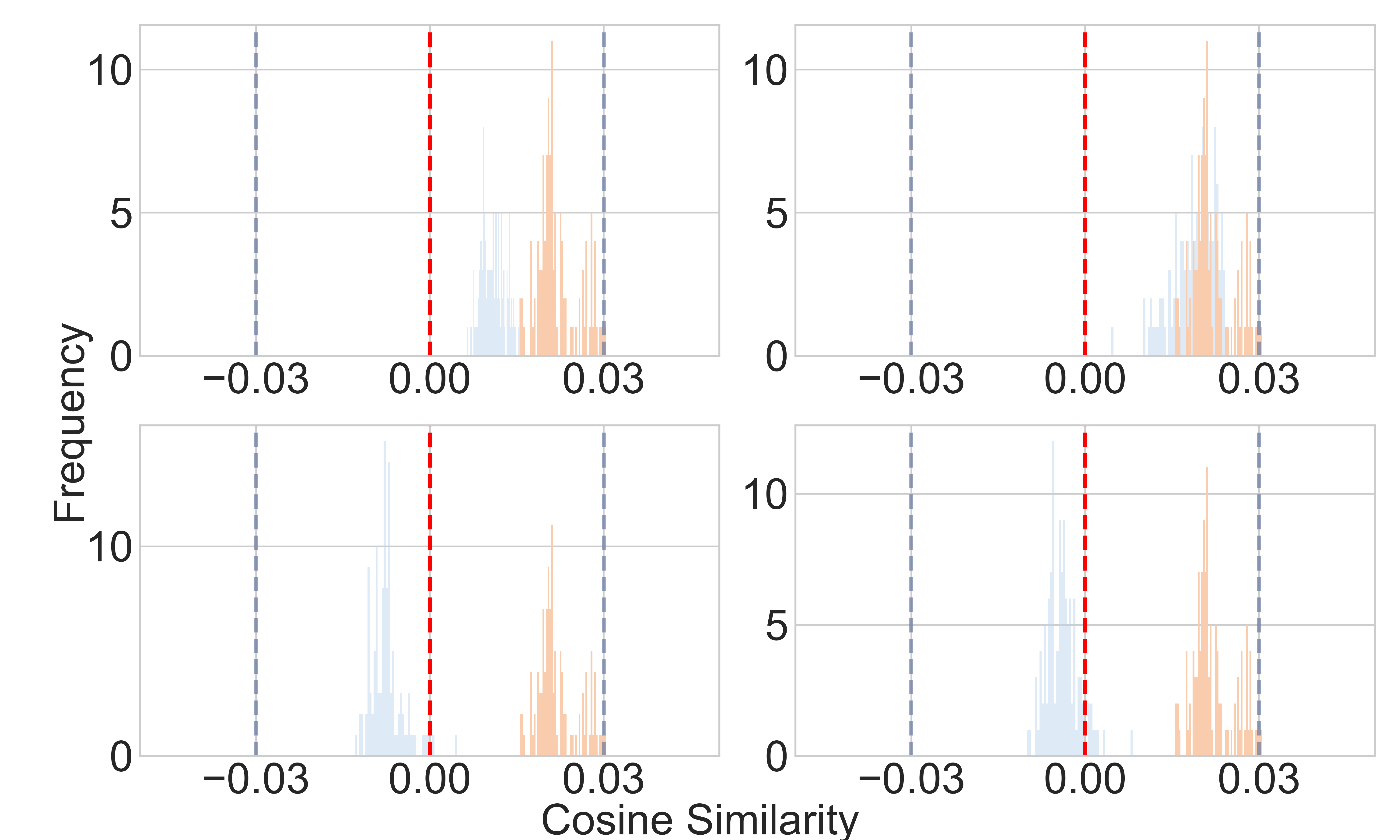}}
	\caption{The histograms of cosine similarities between angles between token and the gating weight.}
	\label{freq_2}
\end{figure}

To verify our proof on Lemma \ref{lma:router}, the angle between tokens, as well as the angle between the token and the gating weight, are explored, shown in Figure \ref{freq_1} and \ref{freq_2}, respectively. Figure \ref{freq_1} at row $i$ and conlumn $j$ plots the distribution of cosine similarities between every pair of tokens routed to the expert $i$ and $j$, respectively. Specifically, the diagonal ones denote the distributions of tokens from the same expert. It can be seen that the tokens routed to the same expert are more alike, with the cosine similarity closer to 1.

Then, we select a representative expert to discuss the phenomenon further. Figure \ref{freq_2} illustrates the frequency of cosine similarity between tokens and gating weights. The orange bar stands for the cosine distribution of the angle between the token and the gating weight, which corresponds to the expert where the token is routed. The blue bar denotes the distribution of the cosine similarity between the above tokens and another expert. Obviously, most tokens close to the specific expert are indeed routed to it, and the distribution has wide differences with other experts. From our experiments, the $\delta$ in Formula \ref{thm:delta} is about $0.03$.

\subsection{Ablation Analysis}

The ablation study is built around aspects including the proportion of computation and communication time, load equalization, and astringency. Moreover, in order to prove that the modifications do not affect the model accuracy, the results for inference are also evaluated for verification.

\subsubsection{Proportion of Computation and Communication}

\begin{figure}
	\centerline{\includegraphics[width=8cm]{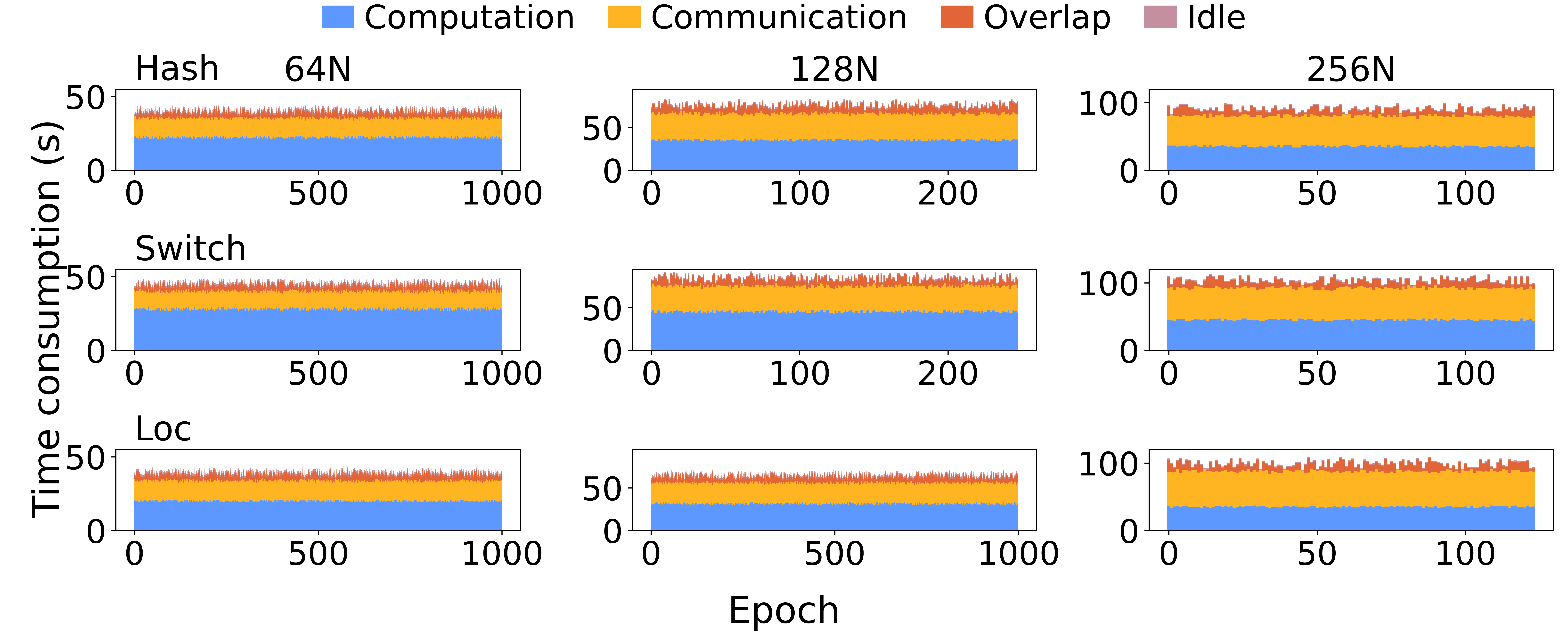}}
	\caption{The composition of training time in each epoch under our cluster groups.}
	\label{time_ratio_epoch}
\end{figure}

We record the total elapsed time per epoch as well as the time consumption for computation, communication, overlapping, and idle with MoEs under different cluster configurations, as shown in Figure \ref{time_ratio_epoch}. Under the model configuration in this paper, each epoch contains 8 steps, and there are 16 experts in total. Following the analysis of the average time consumption per step, LocMoE has both minimal computation overhead and communication overhead under 64N and 128N. However, under 256N, although LocMoE still has the lowest computation costs, its performance does not surpass the HashMoE. The reason is that load balance is more critical than locality when some devices may not have experts. Due to some of the aforementioned engineering optimizations, the propotion of elapsed computation time of LocMoE slightly fluctuates when the amount of devices increases. Meanwhile, the proportion of communication also rises, and the degree of overlapping becomes deeper.

\begin{figure}
	\centerline{\includegraphics[width=8cm]{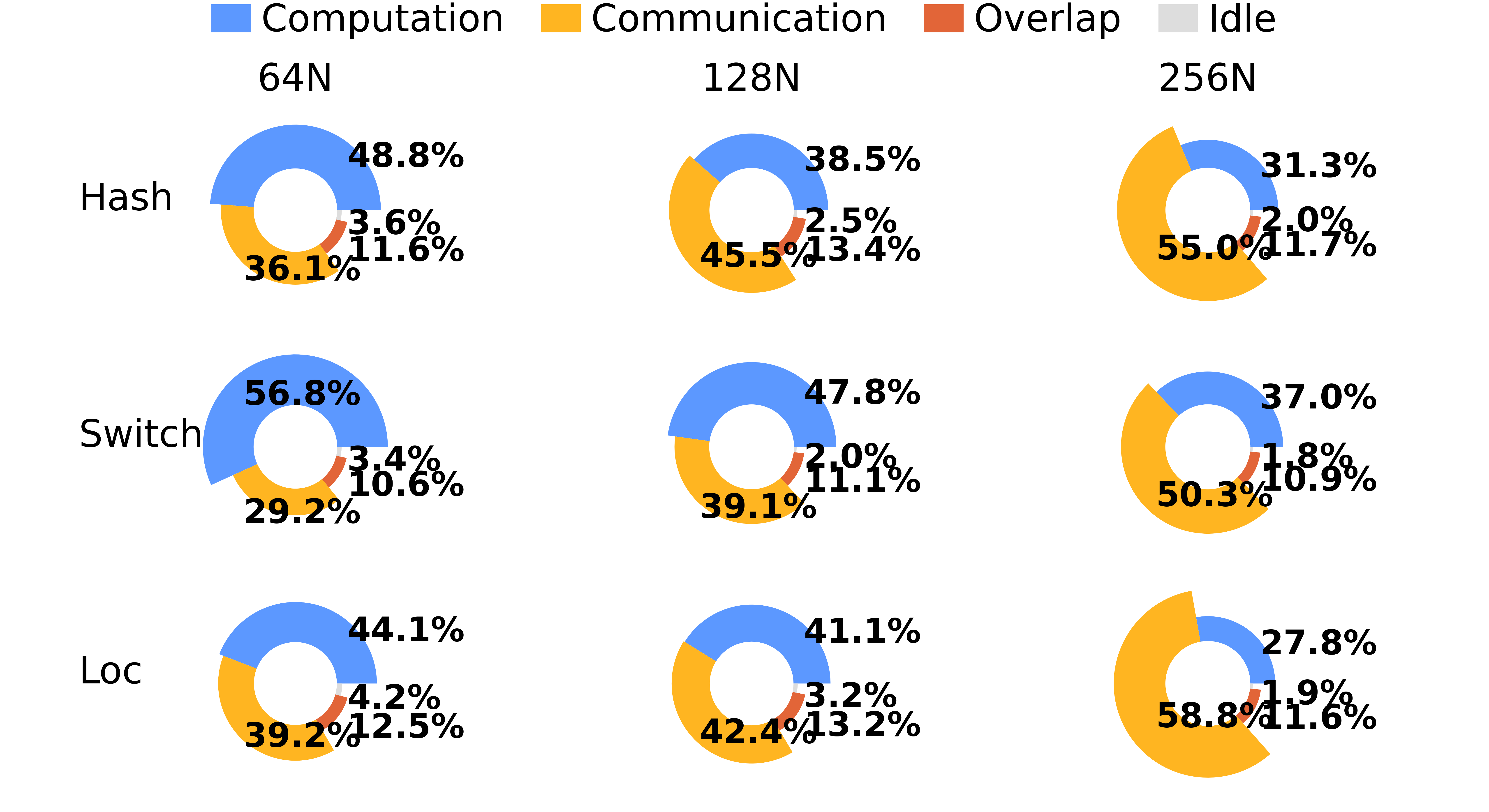}}
	\caption{The time consumption ratio with different MoEs under our cluster groups.}
	\label{time_ratio_overall}
\end{figure}

The overall time consumption proportions are shown in Figure \ref{time_ratio_overall}, and the impact of our innovations on these operations can be visually detected. LocMoE always has a relatively smaller time proportion of computation and a higher overlapping proportion compared to SwitchMoE. The actual computation time of LocMoE approaches or is a bit lower than HashMoE, which has no extra computation for token features. When the resource increases, the time proportions of communication for these MoEs all reflect an increasing trend. Specifically, from 64N to 128N, the increasing communication proportion in LocMoE is not as tangible as HashMoE and SwitchMoE. It is shown that the communication time of LocMoE is markedly elevated under 256N with 32 nodes as was expected. The phenomenon indicates that LocMoE is more appropriate for cases whose number of experts is larger than that of nodes. The locality would lose efficacy when the local expert does not exist. Overall, LocMoE offers more notable enhancements in computation.

\subsubsection{Distribution of Expert Assignment}

\begin{figure*}
	\centerline{\includegraphics[width=16cm]{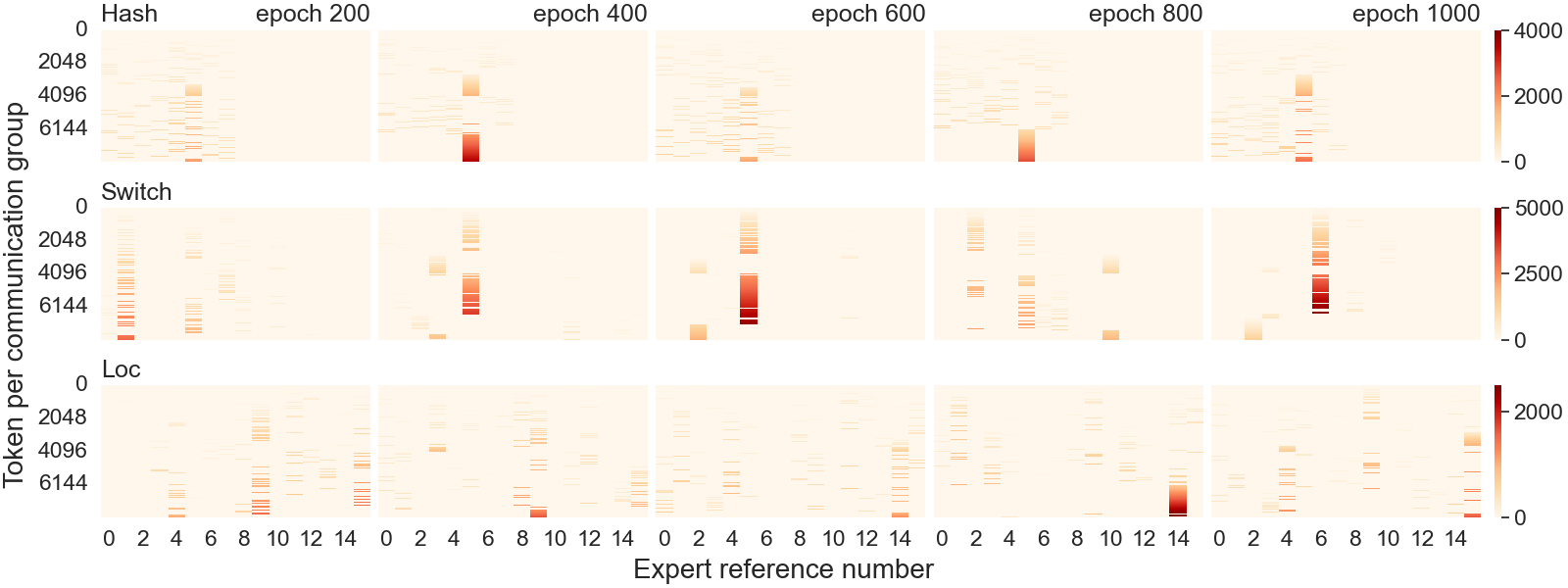}}
	\caption{The allocation of tokens with different routing strategies.}
	\label{expert_position}
\end{figure*}

Taking the cluster of 64N as an instance, Figure \ref{expert_position} portrays the distribution of expert assignments in different MoEs during the training process. Since HashMoE adopts an absolute balance strategy, the allocation of tokens is quite balanced at initialization. However, SwitchMoE and LocMoE initialize from the allocation to a single expert. To avoid misinterpretation, the analysis begins at epoch 200. The vertical axis indicates the number of tokens assigned to each communication group, and the horizontal axis indicates the index of experts. There are 16 experts in our experiments, and the index range is from 0 to 15. The cumulative number of tokens routed to expert $i$ can be observed along the specific vertical axis corresponding to the expert. Each occurrence of a non-zero value means a new token being assigned to this expert. Successive color bars indicate that shuffled tokens are continuously assigned to the same experts, thus causing imbalances to arise. As can be seen from the figure, the rigid constraints in HashMoE ensure that its assignment is even. However, almost no token is routed to experts with an index of 9 to 15 in the subfigure of SwitchMoE. It results in nearly 40\% of the experts' invalidation; to make matters worse, the phenomena of "winner-take-all" is pronounced in expert number 5 and expert number 6. LocMoE, due to the localized bootstrapping, can allocate the token to these experts evenly during the training process, indicating that the dual constraints of auxiliary and locality loss can steadily enhance resource utilization.

\subsubsection{Astringency and Accuracy}

\begin{figure}
	\centerline{\includegraphics[width=8cm]{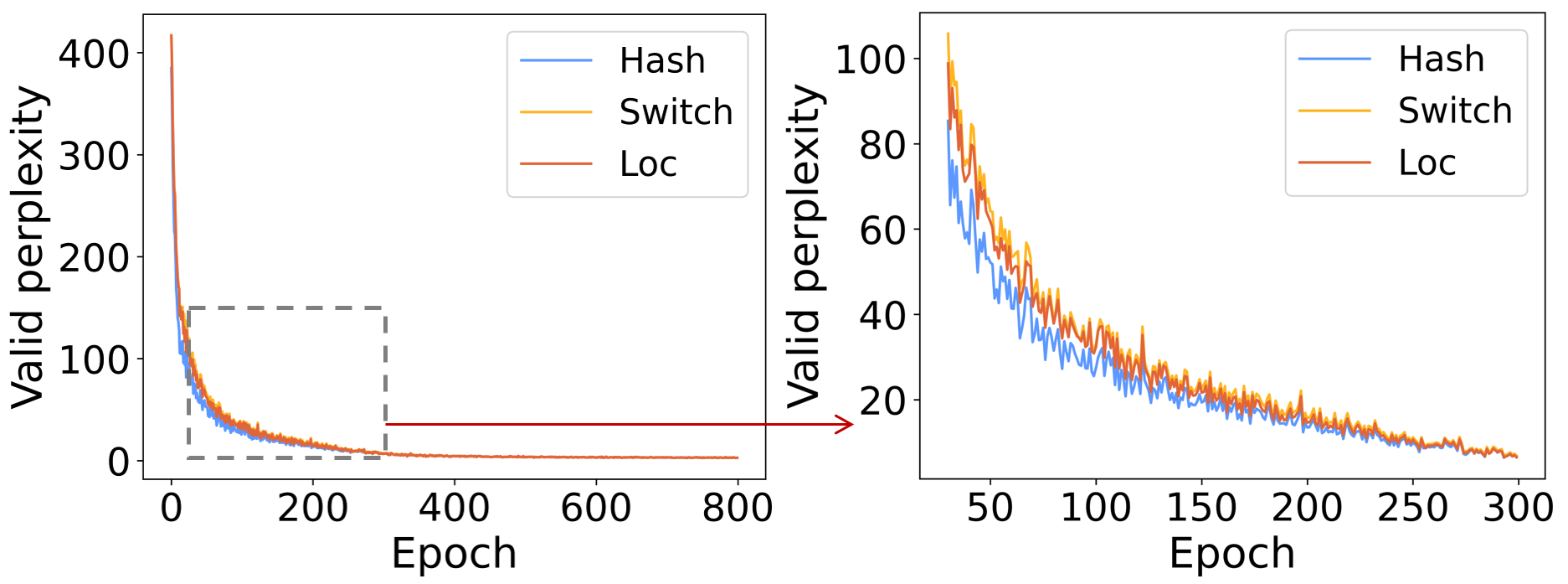}}
	\caption{The valid perplexity throughout the training stage of multiple MoEs.}
	\label{valid_perplexity}
\end{figure}

The astringency is measured by the valid perplexity throughout the process, and the comparison of the convergence speed under different MoEs is depicted in Figure \ref{valid_perplexity}. The overall convergence speed of LocMoE is between that of HashMoE and SwitchMoE in the early stage, and they have an analogical tendency of convergence after a certain amount of epochs. 

HashMoE exhibits better convergence performance due to the fixed and uniformly assignment of RRE. This phenomenon may be perverse because the unlearned routers make it hard to distinguish experts and converge rapidly. The reason for such a situation may be the relatively small angle between tokens in corpora. Concretely, from Appendix B, the dataset contains the fine-grained classification of materials in a specific domain, and the similarities between these items are inherently high. Thus, the composition of the dataset needs to be ameliorated. As for LocMoE, more experts participate in the early training process due to locality. Compared to SwitchMoE, whose routing probability relies only on the token feature, it may promote astringency using LocMoE.

The performance on multiple NLP tasks (see Appendix E) compared with the original PanGu-$\Sigma$ is illustrated in Figure \ref{radar}. All models are pre-trained with the corpora introduced in Appendix B from scratch. The LocMoE and the baseline (original PanGu-$\Sigma$) are both more adept at the query type, while they have difficulties with tasks of the fault tree. The samples of query type are displayed in Appendix F. Due to the enhancement of discrimination for experts and tokens, the comprehension and expressive ability of semantics in various tasks is generally improved.

\begin{figure}
	\centerline{\includegraphics[width=6cm]{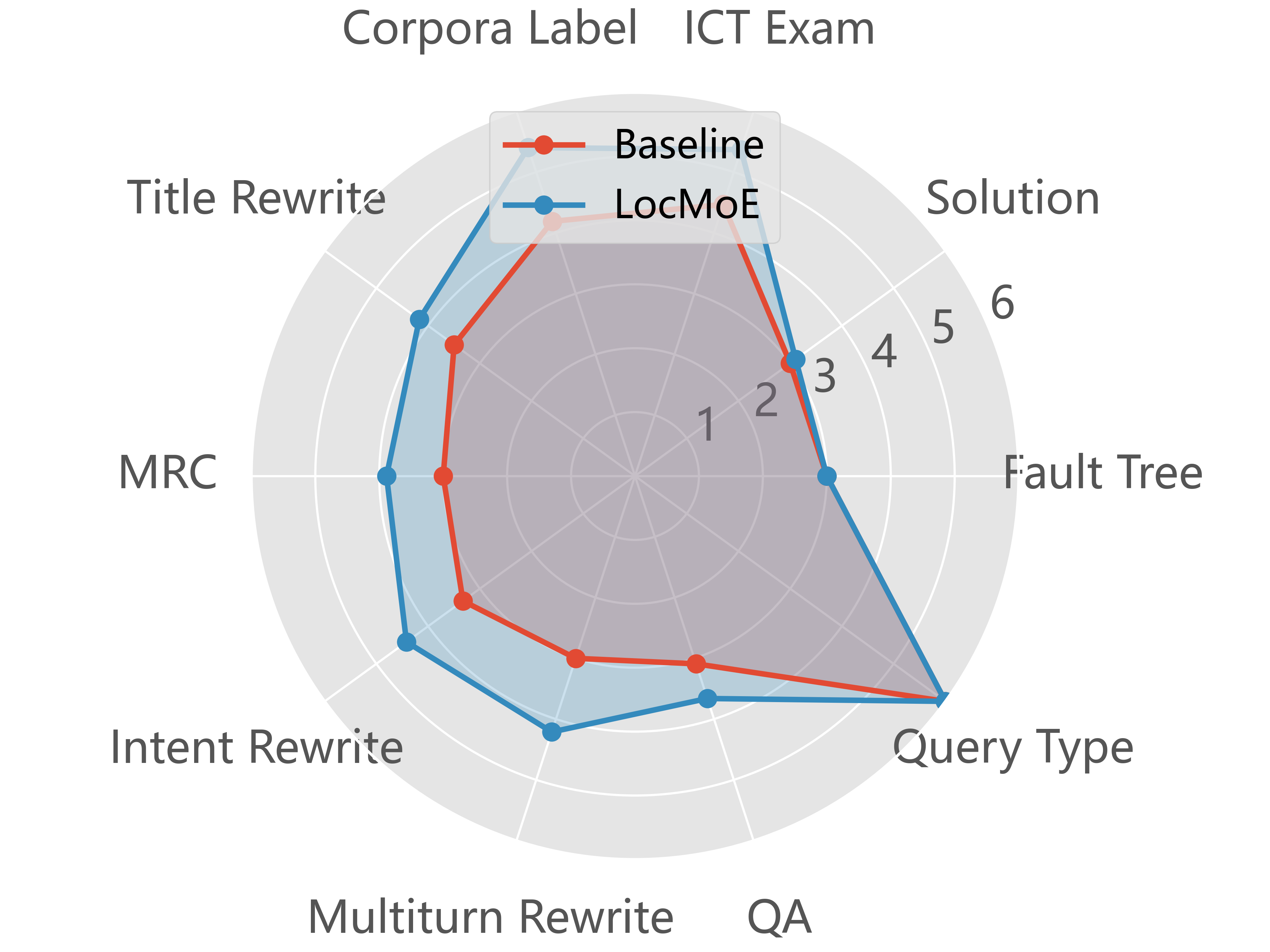}}
	\caption{The scores of inference tasks compared with the baseline.}
	\label{radar}
\end{figure}

\section{Conclusion}

In this paper, we propose a low overhead structure named LocMoE to relieve the performance bottleneck of existing MoE. The modifications mainly revolve around the mechanism of token assignment. The locality loss,  which can be delineated as the distribution difference of token assignments, is proposed to promote locality computation on the premise of load balance. We also provide the theoretical demonstration for the lower bound of the expert capacity to achieve the same effect by training fewer tokens. To meet the assumption of orthogonal gating weight, the GrAP layer is adopted instead of the dense layer to calculate the gating values, and it can also reduce the overhead of computation. Incorporating group-wise All-to-All and communication overlapping features, the elapsed time of communication is further reduced. The experiments are performed on Ascend clusters with 64, 128, and 256 910A NPUs. Compared with current state-of-the-art MoEs, the performance improvement of training is up to 22.24\%. Evaluating multiple NLP tasks, it is detected that the interactive capability of our model is also enhanced. From the results that explore the relationship between the scale of expert capacity and the token features, we find that the dataset construction still needs to be improved. In future work, we will further organize the multilingual corpora from more fields.

\clearpage
\section*{Appendix}
\label{sec:appendix}

\subsection*{A. Proof Sketch in 3.2}

\subsubsection*{A.1 Proof for Lemma 1}

\begin{proof}
	According to the previous definition, $\delta_{i^*, j} = \cos(\theta_{i^*, j})$, where $i^*$ is the expert that the token $j$ routed to. $\theta_{i^*, j}$ is the angle between token $j$ and the gating weight $\omega_{i^*}$ corresponding to the expert $i^*$. Combined with Formula (3) and (4) in Section 3.2, we have:
	
	\begin{equation}
		\begin{aligned}
			i^* &= \mathop{\arg\max}\limits_{i\in{[n]}}(\langle{\omega_i, x_m}\rangle) \\
			\mathrm{where}\;\langle{\omega_i, x_m}\rangle &= \Vert{\omega_i}\Vert{\cdot}\Vert{x_m}\Vert{\cdot}\cos(\theta_{i^*, j}) \\
			i^* &= \mathop{\arg\max}\limits_{i\in{[n]}}(\langle{\omega_i, x_m}\rangle) \\
			&= \mathop{\arg\max}\limits_{i\in{[n]}}(\delta_{i^*, j})
		\end{aligned}
	\end{equation}
\end{proof}

\subsubsection*{A.2 Proof for Lemma 3}

\begin{proof}
	The area of a hyperspherical cap in a $n$-sphere of radius $r$ can be obtained by integrating the surface area of an $(n-1)$-sphere of radius $r \sin \theta$ with arc element $r \mathrm{d} \theta$ over a great circle arc, that is:
	
	\begin{equation}
		\begin{aligned}
			& A_n^{\mathrm{cap}}(r)=\int_0^\phi A_{n-1}(r \sin \theta) r \mathrm{d} \theta \\
			& =\frac{2 \pi^{(n-1) / 2}}{\Gamma\left(\frac{n-1}{2}\right)} r^{n-1} \int_0^\phi \sin ^{n-2} \theta \mathrm{d} \theta \\
			& =\frac{2 \pi^{(n-1) / 2}}{\Gamma\left(\frac{n-1}{2}\right)} r^{n-1} J_{n-2}(\phi) \\
			& =\frac{2 \pi^{(n-1) / 2}}{\Gamma\left(\frac{n-1}{2}\right)} r^{n-1} \frac{1}{2} B\left(\frac{n-1}{2}, \frac{1}{2}\right) I_{\sin ^2 \phi}\left(\frac{n-1}{2}, \frac{1}{2}\right) \\
			& =\frac{1}{2} \frac{2 \pi^{(n-1) / 2}}{\Gamma\left(\frac{n-1}{2}\right)} r^{n-1} \frac{\Gamma\left(\frac{n-1}{2}\right) \Gamma\left(\frac{1}{2}\right)}{\Gamma\left(\frac{n}{2}\right)} I_{\sin ^2 \phi}\left(\frac{n-1}{2}, \frac{1}{2}\right) \\
			& =\frac{1}{2} \frac{2 \pi^{n / 2}}{\Gamma\left(\frac{n}{2}\right)} r^{n-1} I_{\sin ^2 \phi}\left(\frac{n-1}{2}, \frac{1}{2}\right) \\
			& =\frac{1}{2} A_{n}(r) I_{\sin ^2 \phi}\left(\frac{n-1}{2}, \frac{1}{2}\right)
		\end{aligned}
	\end{equation}
	
	where $A_{n}(r)$ denotes the area of the high-dimensional sphere. $p_\delta$ can be viewed as the proportion of the symmetrical areas formed by $\theta$ to that of the entire sphere, shown as Figure \ref{sphere}:
	
	\begin{figure}[htbp]
		\centerline{\includegraphics[width=6cm]{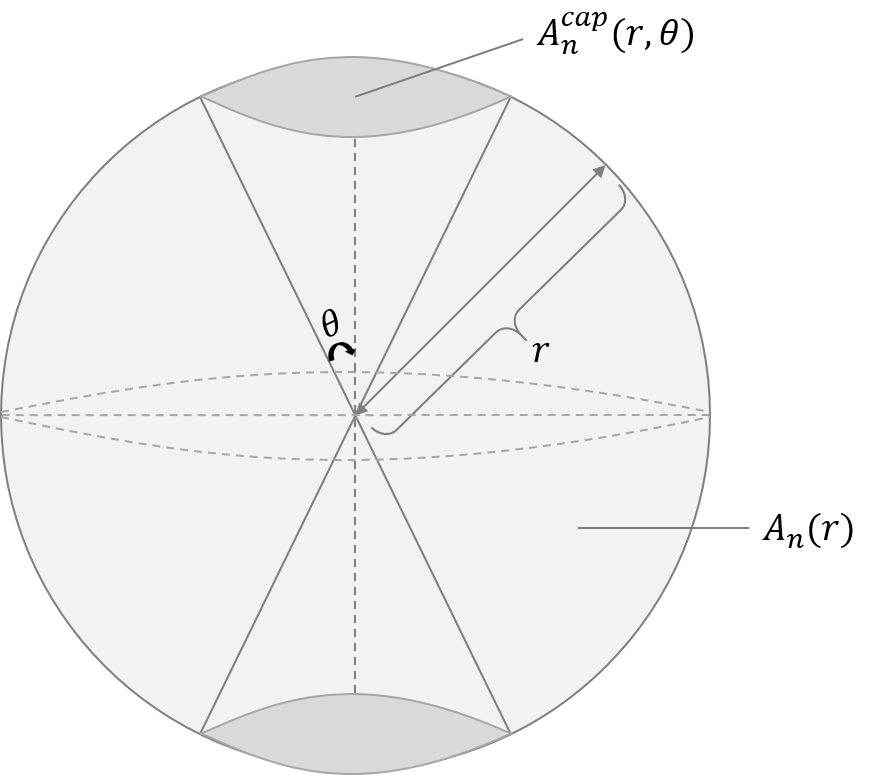}}
		\caption{The schematic of $p_\delta$}
		\label{sphere}
	\end{figure}
	
	\begin{equation}
		\begin{aligned}
			p_{\delta} &= \frac{2{A_n}^{\mathrm{cap}}(r, \theta)}{A_n(r)} \\
			&= I_{1-\delta^2}\left(\frac{d - 1}{2}, \frac{1}{2}\right) \\
			&= 1-I_{\delta^2}\left(\frac{1}{2}, \frac{d - 1}{2}\right)
		\end{aligned}
	\end{equation}
	
	Suppose $\delta = \sqrt{\frac{1}{d - \frac{3}{2}}}$, when $d$ is large, $\delta$ approximates to $\sqrt{\frac{1}{d}}$, then:
	
	\begin{equation}
		\begin{aligned}
			I_{\delta^2}(\frac{1}{2}, \frac{d-1}{2}) &\approx I(\frac{\delta^2(d - 1 + \frac{1}{2} - 1)}{2 - \delta^2}, \frac{1}{2}) + \Theta[(\frac{d-1}{2})^{-2}] \\
			&\approx I(\frac{\frac{1}{d - \frac{3}{2}}(d - \frac{3}{2})}{2 - \frac{1}{d - \frac{3}{2}}}, \frac{1}{2}) \\
			&= I(\frac{1}{2}(\frac{d - \frac{3}{2}}{d - 2}), \frac{1}{2}) \\
			&= \frac{1}{\Gamma{\frac{1}{2}}}\int_{0}^{\frac{1}{2}(\frac{d - \frac{3}{2}}{d - 2})}\exp(-t)t^{\frac{1}{2}}\mathrm{d}t \\
			&\approx \frac{1}{\Gamma(\frac{1}{2})}\int_{0}^{\frac{1}{2}}e^{-t}t^{-\frac{1}{2}}\mathrm{d}t \\
			&= \frac{1}{\Gamma(\frac{1}{2})}\gamma(\frac{1}{2}, \frac{1}{2}) \\
			&= \mathrm{erf}(\frac{\sqrt{2}}{2}) 
		\end{aligned}
	\end{equation}
	
	where $\gamma$ is the incomplete gamma function. Combined with Formula (2) in Section 3.2, $\mathrm{erf}(\frac{\sqrt{2}}{2}) \approx {0.68}$, then:
	
	\begin{equation}
		p_{\delta} = 1 - I_{\delta^2}(\frac{1}{2}, \frac{d - 1}{2}) \approx {0.3}
	\end{equation}
\end{proof}

\begin{table*}
	\centering
	\begin{tabular}{cp{9cm}c}
		\toprule
		\textbf{Hyperparameter} & \textbf{Description} & \textbf{Value}\\
		\midrule
		adam\_eps & Terms to increase the stability of numerical calculations & 1e-6\\
		batch\_size & The size of data input to the model for training each time, related to the number of devices & 32 \\
		expert\_num\_per\_dp\_dim & Number of experts per communication group & 1 \\ 
		expert\_parallel & Number of experts in parallel & 16 \\ 
		moe\_layer\_num & Number of MoE layers & 8  \\ 
		num\_heads & Number of parallel heads & 40  \\
		op\_level\_model\_parallel\_num & Number of parallel models & 8  \\
		sink\_size & The size of data executed per sink & 16  \\
		\bottomrule
	\end{tabular}
	\caption{The critical hyperparameters in configuration of PanGu-$\Sigma$.}
	\label{tab:cfg}
\end{table*}

\subsubsection*{A.3 Proof for Theorem 1}

\begin{proof}
	Refer to the assumption about distributions of class-discriminative and class-irrelevant patterns in pMoE \cite{chowdhury2023patch}, with analogy, the tokens satisfy $\delta_{i, j} \geq {\delta}$ can be regarded as the class-discriminative token. Then, the problem we need to explore can be converted to find the minimum amount of tokens that make at least one class-discriminative token routed to expert $i$.
	
	Suppose $p_i$ is the probability that the token routed to the expert $i$ is a class-discriminative token; we have:
	
	\begin{equation}
		p_i \leq \frac{p_{\delta}}{\frac{1}{n}} = n[1 - I_{\delta^2}(\frac{1}{2}, \frac{d - 1}{2})]
	\end{equation}
	
	where the first inequality holds since the token satisfies $\delta_{i, j} \geq {\delta}$ may not always be routed to expert $i$. Then, the minimum value of expert capacity under the circumstance of at least one class-discriminative token routed to expert $i$ can be written as:
	
	\begin{equation}
		\begin{aligned}
			ec_{\min} = \frac{1}{p_\mathrm{s}} &= \frac{1}{n[1 - I_{\delta^2}(\frac{1}{2}, \frac{d - 1}{2})]} \\
			\mathrm{For\;large\;}d,\;I_{\delta^2}(\frac{1}{2}, \frac{d - 1}{2}) &\approx I(\frac{\delta^2(d - \frac{3}{2})}{2 - \delta^2}, \frac{1}{2}) \\
			& \approx \frac{1}{\Gamma(\frac{1}{2})}\gamma(\frac{1}{2}, \frac{\delta^2d}{2-\delta^2}) \\
			& = 1 - \mathrm{erfc}(\sqrt{\frac{\delta^2d}{2-\delta^2}}) \\
			\mathrm{thus},\;ec_{\min} & \geq \frac{1}{n\cdot{\mathrm{erfc}(\sqrt{\frac{\delta^2d}{2-\delta^2}})}} \\
			& > \frac{1}{n}\exp(\frac{\delta^2d}{2-\delta^2})
		\end{aligned}
	\end{equation}
\end{proof}

\subsection*{B. Datasets}

PanGu-$\Sigma$ has already demonstrated its ability to learn efficiently and independently from text corpus in various domains. In this work, we will evaluate the performance of PanGu-$\Sigma$ in detailed knowledge of a specific area. The materials connected to mobile network operators' services are chosen as input corpora. Concretely, blogs and technical documents in the form of \emph{iCase}, \emph{Wiki}, core network/Man-Machine language (MML), configuration translations, feature documents, etc., are collected. These corpora are in Chinese, English, or bilingual (Chinese-English). 

Among them, \emph{iCase} indicates the technology case, which records procedures of problem handling and contributes to problem delimitation and localization. \emph{iCase} contents include the wireless network, optical, carrier IT, cloud core network, network energy, etc. It contains code of Java, SQL, Shell, other programming languages or commands, and the related logs, totaling 591,972 documents (368,282 Chinese, 223,690 English, 1.7GB) and 387,223,874 tokens. \emph{Wiki} is the document extracted from 3ms (Huawei's internal knowledge management platform). Topics of Wiki include insight reports, R\&D tool guides, training summaries, industry standards, configuration manuals, etc., totaling 1,146,755 documents (1,118,669 in Chinese, 27,632 in English, and 454 bilingual, 4.1 GB) and 116,152,3537 tokens. The corpora in the field of core network and MML are mainly derived from the product information from mobile network operators or public platforms, such as 3GPP protocols, customized specifications, high-quality MO Support Processes (MOP), engineering solutions, and MML scripts for existing networks, totaling 223,898 documents (all in Chinese, 0.476GB) and 136908105 tokens. Configuration translation data come from product documents for data communication equipment of Huawei or Cisco involving switches, firewalls, and routers, totaling 1460680 documents (all in Chinese, 2.2 GB) and 559716720 tokens. Feature documents include product design documents for data communication, IT and other business lines, 4G/5G feature documents, the frequently asked question (FAQ) of machine question and answering (Q\&A), fault trees, fault location guides, etc., totaling 86,913 documents (52,677 in Chinese, 34,236 in English, 0.29GB).

The above corpora are in different formats: Word, PDF, HDX, and HTML. First of all, the original corpora need to be parsed. For instance, The text of a PDF document is extracted with the pattern recognition technique, and the machine Q\&A corpus is manually entered by iCare engineers. After that, the fine-grained corpora are merged and organized into a complete sample to ensure a complete thought chain. Taking MML scripts as an example, their structuredness is divided into three levels from global to local: (1) Features composed of medium features; (2) Medium features composed of multiple ordered MMLs; (3) MML instances. Product documents can uniquely identify medium features, and the diversity of MML instances can be constructed from the present network's MMLs. The corpora are refined; that is, after removing meaningless symbols and descriptions, duplication elimination is performed on the corpora based on text similarity and semantics to avoid overlapping data. The next step is to regularize the data, including removing private data and unifying the specification of forms and process symbols. Finally, a customized tokenizer based on the domain dictionary is applied to the participle, and the cleaned corpora are obtained for training.

\subsection*{C. Experimental Environment}

The experiments are conducted on Ascend clusters, and the environment falls into three groups: 64, 128, and 256 Ascend 910A NPUs. The Ascend 910A series NPU has 32 AI Cores, with a maximum memory capacity of 2TB and a maximum memory bandwidth of 1.07TB/s. The collective communication function on high-speed links such as PCI-E, HCCS, and RoCE is realized by HCCL, a high-performance collective communication library based on the Ascend. It provides communication primitives on single-node-multi-card and multi-node-multi-card, and it also supports various communication algorithms such as ring, mesh, HD, ring + HD, and mesh + HD.

The versions of the Compute Architecture for Neural Networks (CANN) suite (toolkit, CANN, driver) are 5.1.RC2.1, 1.84, and 23.0.rc2, respectively. The CANN is the heterogeneous computing architecture developed by Huawei, and it supports multiple AI frameworks, including MindSpore, PyTorch, TensorFlow, etc., providing interfaces to build AI applications on the Ascend platform. Our model runs on the MindSpore framework with version 2.0.0.

\subsection*{D. Model Configuration}

The hyperparameter configuration of our model is listed in Table \ref{tab:cfg}. Thereinto, \emph{batch\_size} and \emph{sink\_size} are relevant to the number of devices, and the values in the table are under 128N. The total number of experts can be obtained by \emph{expert\_num\_per\_dp\_dim} * \emph{expert\_parallel}.

\subsection*{E. Measurement Metric}

We design multiple NLP tasks to systematically evaluate the knowledge understanding and semantic expression capabilities of our model. These tasks are extracted from 10 business perspectives in the field of carrier networks, such as fault tree nodes, solutions, ICT certification exams, and title rewriting. Among them, taking the recognition fault tree node as an example, the construction of the NLP task is divided into two steps: firstly, the text is differentiated into difficulty levels (L1 to L3) according to the logical complexity of concepts and inter-conceptual relationships, and the samples are selected according to the hierarchies. L1 contains single and a group of connection parameters with integrity and independence, complete temporal connection parameters; L2 represents quantitative relationship parameters, referential relationship parameters, and combination parameters; L3 denotes the sample that cannot be intellectualized. In the next step, after the classification is completed, the prompt, derived from the structured specification of the fault discriminative approach, is applied to generate structured parameters and restores the discrimination logic.

Q\&A pairs are organized for each task, and 30 to 80 items from among them are picked off as the evaluation set. The original PanGu-$\Sigma$ that goes through the same pre-training process acts as the baseline; then, the review task is input individually to get answers. Staff in DataLab are invited to grade manually on the quality of these answers. Ultimately, the average scores for each task are recorded with removing discrete values.

\section*{Acknowledgments}

We thank all anonymous reviewers for their valuable feedback. We would like to express our appreciation to Dachao Lin (Dr. Lin) for the improvements in theoretical proof in this paper. We are grateful for the guidance from the Noah's Ark Lab on model training. Moreover, the contributions of the MindSpore team and employees participating in model evaluation are greatly appreciated.

\section*{Contribution Statement}

Jing Li, Zhijie Sun, and Xuan He wrote this paper and contributed equally to this work. Li Zeng, Yi Lin, Entong Li, and Binfan Zheng provided experimental analysis and offered suggestions for this paper. Rongqian Zhao and Xin Chen are the project leaders and provide support for this work.

\bibliographystyle{named}
\bibliography{custom}

\appendix

\end{document}